\documentclass[a4paper,fleqn]{cas-dc}

\usepackage[authoryear,longnamesfirst]{natbib}
\usepackage{multirow}
\def\tsc#1{\csdef{#1}{\textsc{\lowercase{#1}}\xspace}}
\tsc{WGM}
\tsc{QE}
\tsc{EP}
\tsc{PMS}
\tsc{BEC}
\tsc{DE}

\begin{document}
\let\WriteBookmarks\relax
\def\floatpagepagefraction{1}
\def\textpagefraction{.001}

\shorttitle{RLID: Defending Adversarial Patches}

\shortauthors{Junwen Chen, Xingxing Wei et~al.}

\title [mode = title]{Defending Adversarial Patches via Joint Region Localizing and Inpainting }                      



%
\author[1]{Junwen Chen}[style=chinese,
                        orcid = 0000-0003-1461-4452 ]



\ead{zy2106122@buaa.edu.cn}


\credit{Conceptualization,  Methodology, Writing – original draft } 

\affiliation[1]{organization={School of Computer Science and Engineering, Beihang University},
    addressline={No.37, Xueyuan Road, Haidian District}, 
    country={China}}



\author[2]{Xingxing Wei}[style=chinese]

\cormark[1]


\ead{xxwei@buaa.edu.cn}


\credit{Conceptualization of this study, Writing – review and editing }

\affiliation[2]{organization={Institute of Artificial Intelligence, Beihang University},
    addressline={No.37, Xueyuan Road, Haidian District}, 
    country={China}}
\cortext[cor1]{Corresponding author}



\begin{abstract}
 Deep neural networks are successfully used in various applications, but show their vulnerability to adversarial examples. With the development of adversarial patches, the feasibility of attacks in physical scenes increases, and the defenses against patch attacks are urgently needed. However, defending such adversarial patch attacks is still an unsolved problem. In this paper, we analyse the properties of adversarial patches, and find that: on the one hand, adversarial patches will lead to the appearance or contextual inconsistency in the target objects; on the other hand, the patch region will show abnormal changes on the high-level feature maps of the objects extracted by a backbone network. Considering the above two points, we propose a novel defense method based on a ``localizing and inpainting" mechanism to pre-process the input examples. Specifically, we design an unified framework, where the ``localizing" sub-network utilizes a two-branch structure to represent the above two aspects to accurately detect the adversarial patch region in the image. For the ``inpainting" sub-network, it utilizes the surrounding contextual cues to recover the original content covered by the adversarial patch. The quality of inpainted images is also evaluated by measuring the appearance consistency and the effects of adversarial attacks. These two sub-networks are then jointly trained via an iterative optimization manner. In this way, the ``localizing" and ``inpainting" modules can interact closely with each other, and thus learn a better solution. A series of experiments versus traffic sign classification and detection tasks are conducted to defend against various adversarial patch attacks. The results show that our defense method has achieved an increase in the accuracy of 57\% compared with no defense accuracy on average, outperforming the existing state-of-the-art defenses of 37\%, which verifies the effectiveness and superiority of the proposed method.

\end{abstract}







\maketitle

\section{Introduction}
Deep neural network models achieve excellent performance on a variety of tasks such as face detection \cite{Parkhi2015DeepFR}, malware detection \cite{ronen2018microsoft} and autonomous driving \cite{huval2015empirical}. However, recent research works have shown that these models are vulnerable to small perturbations and these so-called adversarial perturbations can result in the  misclassification when the image or video is predicted \cite{akhtar2018threat,goodfellow2014explaining,szegedy2013intriguing,9531416}. Many prior attacks focus on setting global perturbations to an object \cite{wei2019sparse,gao2021push,zhang2022transferable}, which are not always practical in the real world. 
Recently, adversarial patches are intensively studied, in which the adversary adds adversarial perturbations only within a local region, and do not restrict the magnitude of the perturbations. These attacks can be used in the physical world by sticking the adversarial patch to the object, which leads to the  {Deep Neural Networks (DNN)} model’s misclassification and object disappearance \cite{lee2019physical}. Once a strong adversarial patch is generated, it can be shared and applied to some other similar scenes. In that case, adversarial patch attacks effectively against DNN systems using the same models and achieve real-world attacking robustness. Such methods can be successfully and widely used to attack real-world machine learning tasks like image classification \cite{brown2017adversarial,karmon2018lavan} and object detection \cite{DBLP:journals/corr/abs-1909-04326,song2018physical,DBLP:journals/corr/abs-1910-11099}. The success of adversarial patches may cause serious safety problems, so defenses against such attacks are urgently needed.

\begin{figure}
  \centering
  \includegraphics[width=0.95\linewidth]{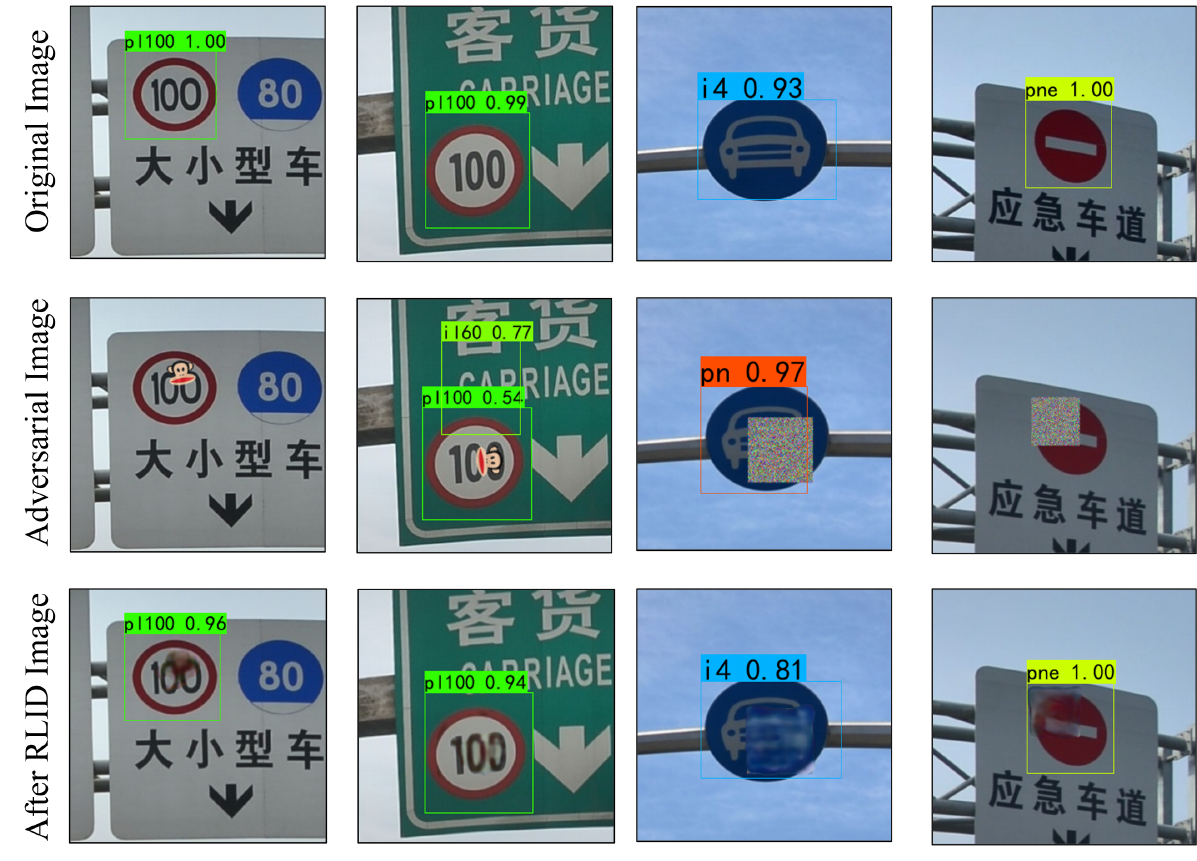}
  \caption{Examples output by RLID method. The top row lists clean examples that can be detected correctly. The second row lists adversarial examples generated by different adversarial attacks. These images mislead the detector via wrong localization or classification results. The third row shows the recovered images  pre-processed by our RLID. And they can be correctly detected again. All images are from TT100k dataset \cite{Zhe_2016_CVPR}. }
  \label{fig:show}
\end{figure}

There are some existing defense methods that are specially designed against adversarial patches, such as Local Gradients Smoothing (LGS) \cite{naseer2019local}, Digital Watermarking (DW) \cite{hayes2018visible}, Segment and Complete defense (SAC) \cite{liu2021segment}, TaintRadar \cite{li2021detecting}, and so on. They all belong to the pre-processing modules, and exploit a ``detecting and removing" strategy. Specifically, the adversarial patch regions are first detected by the pre-defined  criteria or learned segmenter, and then these detected regions are removed from the original image (i.e., replaced with the black pixels). Although the ``removing" operation can eliminate the attacking effect of the adversarial patches, they destroy the contextual information within the original image.  {Therefore, these incomplete images will affect the performance of downstream tasks like image classification or object detection.} Compared with this ``detecting and removing" strategy, a more reasonable option should be ``\textbf{detecting and recovering}" strategy, i.e., recovering the original content covered by the adversarial patch after detecting the adversarial patch region. If the adversarial patch regions are perfectly recovered, the performance in the downstream tasks will not decrease caused by adversarial patch attacks. 

However, ``detecting and recovering" does not mean a simple separate two-phase framework. Because patch's detecting and recovering can interact closely with each other. By utilizing the mutual relationship,  we can learn a global optimal solution to defend adversarial patch. Considering this point, an unified framework with jointly detecting and recovering for adversarial patch is needed. 

In this paper, we propose an effective defense framework based on the above ``detecting and recovering" strategy, which is called Region Localizing and Inpainting Defense (RLID). We utilize region localizing to detect adversarial patches, and use region inpainting mechanism to perform the region recovering. To meet this goal, we analyse the properties of adversarial patches, and find that: on the one hand, adversarial patches will lead to the appearance or contextual inconsistency in the target objects; on the other hand, the patch region will show abnormal changes on the high-level feature maps of the objects extracted by a backbone network \cite{yu2021defending}. Based on these two points, two modules in RLID are carefully designed: Patch Region Localizing (PRL) and Patch Region Inpainting (PRI). \textbf{For PRL module}, we use a two-branch structure to detect the patch area: out of the appearance inconsistency, we identify the edge of the patch; out of the abnormally high values caused by the adversarial patches, we generate a regional prediction by the salient object detection method. Combining the two cues, we achieve precise patch region localization. \textbf{For PRI module},  the surrounding contextual cues are utilized to recover the patch region. Specifically, we use the information of the neighbour area to reconstruct the predicted patch region and initialize the predicted patch region through the basic mean-variance operation combined with the inpainted sub-network for further completion. To measure the quality of the recovered image, we make efforts to check the pixel-level consistency and eliminate the adversarial perturbations. The former is achieved by checking the error with ground-truth images, and the latter is to check the error of high-level feature maps with ground-truth images. During the training process, RLID jointly trains the PRL module and PRI module via an iterative optimization manner. In this way, the ``localizing" and ``inpainting" modules can interact closely with each other, and thus learn a better solution. 
In summary, this paper has the following contributions:

\begin{itemize}
\item We propose RLID method, a novel defense framework against adversarial patch attacks in real-world scenes. Our method is based on a ``localizing and inpainting" mechanism. It first detects the patch's location and then uses the contextual cues to complete the image.

\item We combine the consideration of pixel-level consistency and adversarial consistency to reconstruct the adversarial patch region. For the sake of these two levels, we use a two-branch structure to localize the patch region and use the loss function based on high-level feature maps to eliminate the adversarial perturbations.

\item We conduct a series of experiments to implement state-of-the-art adversarial patch attacks and state-of-the-art defenses to these attacks. We achieve the recovering success rate of 57\% on average, outperforming the existing SOTA defenses of 37\%.  {The reconstructed images also have high picture quality. }
\end{itemize}

\par This paper is organized as follows. Section 2 briefly reviews the related works. We introduce the details of the proposed method in Section 3. Section 4 and Section 5 shows a series of experimental results. Finally, we conclude the whole paper in Section 6.

\section{Related Works}


\subsection{Adversarial Patch Attacks}
Deep neural networks show vulnerability to adversarial attacks, where the classification or detection outputs of the DNNs can be lead to wrong results by adding small perturbations onto the input examples. Prior attack methods evaluate each image pixel and add extra values to some of these pixels to generate adversarial examples. L-BFGS \cite{szegedy2013intriguing}, FGSM \cite{goodfellow2014explaining}, DeepFool\cite{moosavi2016deepfool}, CW \cite{carlini2017towards}, PGD \cite{madry2017towards}, EAD \cite{chen2018ead} and JSMA \cite{papernot2016limitations} are such examples that effectively achieve attacks. Although these methods can generate effective adversarial examples by calculation in digital environments, the attack is not practical in physical scenes. In real scenes, large-area and scattered perturbations are difficult to be imposed on the target object. Meanwhile, the optical equipment used in the detection makes it difficult for the carefully calculated perturbations to achieve the expected attack effect.

The adversarial patch attack creates a scene-independent physical-world attack among the rest. The attack aims to be printed out, placed on the object, and recaptured by a camera to mislead the model \cite{kurakin2016adversarial}. Brown \emph{et al.} \cite{brown2017adversarial} introduce the adversarial patch attack that finds a universal adversarial patch that can be applied to any image and causes the misclassification. At the same time, Karmon \emph{et al.} \cite{karmon2018lavan} introduce an attack called Lavan, which creates localized and visible adversarial perturbations. Liu \emph{et al.} \cite{10.1609/aaai.v33i01.33011028} propose PS-GAN that can simultaneously enhance the visual fidelity and the attacking ability for the adversarial patch. Later, numerous patch attacks are also proposed: Wei \emph{et al.} \cite{wei2022adversarial} conduct the attack in the black-box setting and use realistic meaningful stickers existing in life. This attack utilizes the inbreeding strategy based on a regional aggregation of effective solutions and the adaptive adjustment strategy of evaluation criteria. Mark \emph{et al.} \cite{lee2019physical} design adversarial patches that can be placed anywhere in the image, causing all existing objects in the image to be missed entirely by the detector. Zhao \emph{et al.} \cite{zhao2020object} focus on the adversarial attack on some state-of-the-art object detection models and propose a heatmap-based algorithm and consensus-based algorithm to achieve the attack.

\subsection{Adversarial Defenses}

\subsubsection{Defense Against Adversarial Perturbations}
With the rapid development of attacks, especially with the improvement of the feasibility of the attack in the physical scenes, various defense methods are proposed to eliminate the effect of the adversarial attacks on the model. Previous defense methods are designed in four main directions: 1) Adversarial training. Such defense adds adversarial examples during the training process and achieves higher accuracy during the testing process \cite{zheng2016improving}; 2) Modifying networks's architecture. By adding extra layers or sub-networks and modifying the activation function to enhance the defense ability of the network \cite{buckman2018thermometer}; 3) Using external models as network add-ons. When classifying unseen examples, these methods also input the examples to the external models to help the classifier classify the examples. \cite{samangouei2018defense,song2017pixeldefend}; 4) Preprocessing the adversarial examples. Similar to add-on models, such methods randomize the image to reduce its adversarial attack effect \cite{xie2017mitigating}. Some defense methods may contain two more directions. In general, there are various methods designed to defend against the adversarial perturbations distributed over the entire image.

\subsubsection{Defense Against Adversarial Patch Attacks}

Although there are a large number of defense methods against adversarial attacks, they are not efficient enough to adversarial patch attacks. Among them, adversarial training needs to retrain the entire network each time a new adversarial sample appears, which is not realistic in practical applications; other defenses that use global gradient calculations or denoise based on Gaussian noise are not efficient enough or not suitable for localized patch attacks \cite{wei2022simultaneously} and non-Gaussian patch attacks (e.g., meaning realistic stickers \cite{wei2022adversarial}). So, some defenses against adversarial patch attacks are proposed: Digital Watermarking (DW) \cite{hayes2018visible} is based on the saliency map and uses a combination of erosion and dilation to remove small holes. Local Gradient Smoothing (LGS) \cite{naseer2019local} regularizes gradients in the estimated noisy region before feeding the image to DNN. PatchGuard \cite{xiang2021patchguard} uses small receptive fields to impose a bound on the number of features corrupted by an adversarial patch. Feature Norm Clipping (FNC) \cite{yu2021defending} clips feature norms according to mathematical analysis. Certified defenses \cite{chiang2020certified} is the first certified defense against patch attacks. Randomized Smoothing \cite{levine2020randomized} is de-randomized and deterministic certificates that improves yielding. TaintRadar \cite{li2021detecting} detects localized adversarial examples via analyzing critical regions that attackers have manipulated.

DW method uses a saliency map to detect the adversarial area, which is not always accurate. FNC and PatchGuard need to obtain the network structure. These methods can achieve defensive effects but rely heavily on the structure of the subsequent classifiers or object detectors. Once the subsequent model changes, these defensive models need to be retrained, which means they are not convenient and universal enough. Furthermore, these methods often reduce the useful  class evidence of the true class of the image by performing gradient cropping or smoothing operations on the image, which will result in the loss of the image information. 

In this paper, we propose a novel method to defend the above adversarial patch attacks via a region localizing and inpainting mechanism. The adversarial patch region is first localized, and then is inpainted using the contextual cues around the patch region. These two modules are jointly trained to learn a better solution. 

\begin{figure*}[t]
\centering
\includegraphics[width=0.99\textwidth]{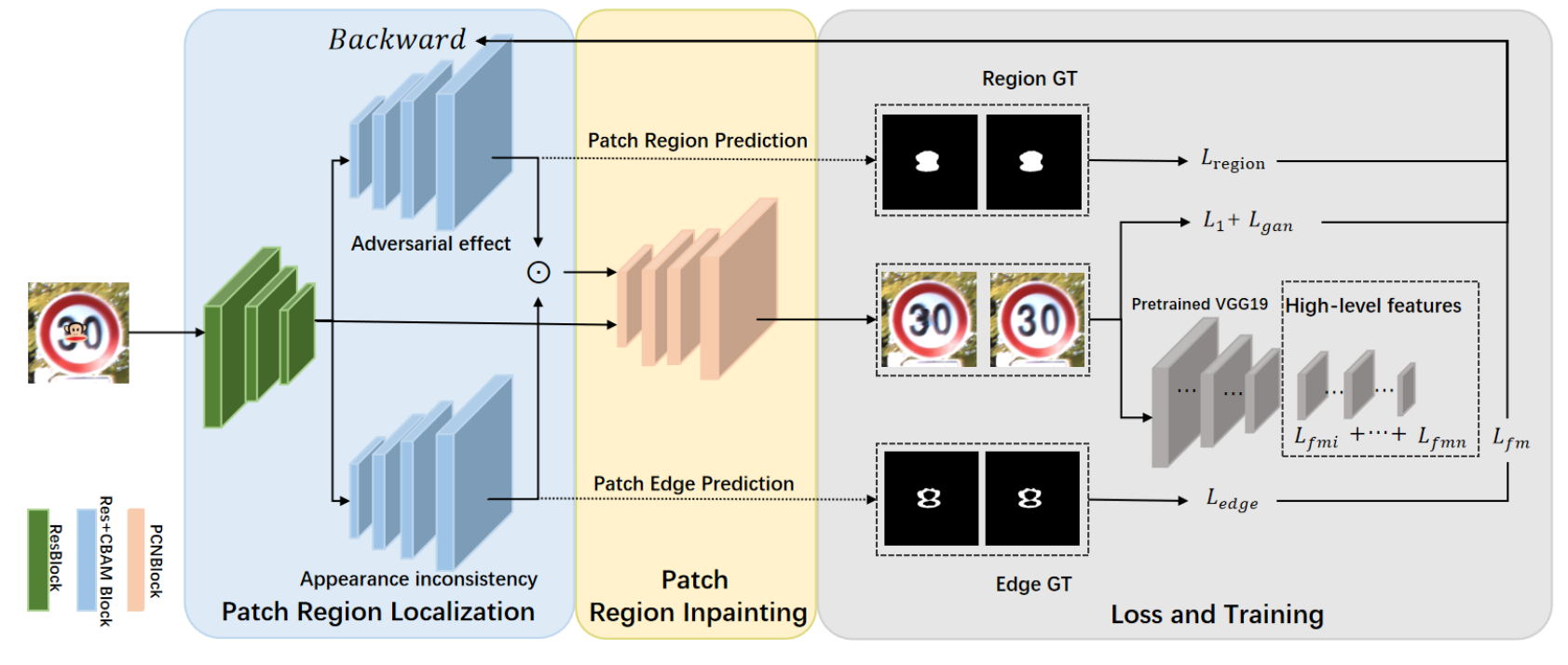}
\caption{\small{The framework of our RLID method. We use two sub-models (the blue part represents Patch Region Localizing (PRL) network and the yellow part represents the Patch Region Inpainting (PRI) network) and the PRL module performers as the two-branch structure by using two decoder structures. The outputs of PRL will guide the inpainting of the PRI module, and the optimizer uses high-level feature maps to eliminate the adversarial perturbations. The image is from GTSRB dataset \cite{STALLKAMP2012323}. }}
\label{fig:framework}
\end{figure*}
\section{Methodology}

\subsection{Adversarial Patch Attacks}\label{sec:criteria}

Adversarial images are created by adding perturbations to images, and the perturbations are pixel-constructed noises or realistic meaningful patterns existing in our life. For an input image $\mathbf{x} \in \mathbb{R}^{c\times h \times w}$, let ${Pr(y \mid \mathbf{x})}$ denote the probability that the model predicts an image $\mathbf{x}$ as the label $\emph{y}$. The former strategies (e.g., AdvP and Lavan) generate an adversarial example as follows: First, generate a random patch's perturbation and a random location. Then, apply a random translation like scaling or rotating on the patch in each image. Finally, use gradient descent to generate the patch over the training set. After obtaining the final patch, an adversary can apply the patch on the image to achieve the attack.

In particular, for a given image  $\mathbf{x} \in \mathbb{R}^{c \times h \times w}$, an initial patch $\emph{p}$, a patch location $\emph{l}$, a target label $\hat{y}$, and patch transformations $\emph{t}$, an adversary defines a patch application operator $\emph{A(p, x, l, t)}$ which first applies the transformations $\emph{t}$ to the patch $\emph{p}$, and then applies the transformed patch $\emph{p}$ to the image $\emph{x}$ at the location $\emph{l}$. The patch is  optimized by solving the following objective function:
\begin{equation}
\hat{p}=\underset{p}{\arg \max } \mathbb{E}_{x \sim X, t \sim T, l \sim L}[\log \operatorname{\emph{Pr}}(\hat{y} \mid A(p, x, l, t)], \label{1}
\end{equation}
where $X$ is the training set of images, $T$ is a distribution over transformations of the patch, and $L$ is a distribution over locations in the image. The targeted attack sometimes uses a selected image from the target class as the initial patch for training to obtain the adversarial patch with some semantic information. Later attack \cite{wei2022adversarial}, instead of generating patches or using another image as the initial patch, uses realistic meaningful stickers to place on the image and achieves its attack effect by computing out the proper position. Obviously, not all stickers can be used as an adversarial patch, so such stickers need to be carefully selected.

\subsection{Our Method}
\label{sec:our_method}

For patch attacks in the physical scenes, we aim to eliminate the possible adversarial noises from examples by processing the inputs from the perspectives of appearance inconsistency and adversarial attack effect. The input processed by our method can enable the subsequent classification model or detection model to make correct judgments. Thus, we can achieve a universal defense against adversarial patch attacks.

Our method is constructed of Patch Region Localizing (PRL) and Patch Region Inpainting (PRI) sub-modules. The main structure of these two module are decoders which input the same output from an encoder. The PRL module aims to localize possible patch areas, while the PRI module performs semantic recovery and adversarial perturbations elimination of the located possible patch areas by relevant semantic areas. These two sub-modules are described in detail below. The specific layer structure can be checked in Table \ref{tab:PRL_REGION} and \ref{tab:PRI}.

\subsubsection{Patch Region Localizing}
\label{subsec:prl}

In previous attack methods, an adversary attacks all pixels of the whole image, while patch attacks only focus on a local region. We define a mask $\bf{M}$ to label the patch region by one and the others by zero. We denote the adversarial example $\mathbf{x}_{adv}$ as:
\begin{equation}
\mathbf{x}_{adv}=(\mathbf{1}-\mathbf{M}) \odot\mathbf{x}+\mathbf{M} \odot\mathbf{\overline{x}},\label{2}
\end{equation}
where $\mathbf{x}$ denotes the clean example, $\mathbf{\overline{x}}$ denotes the adversarial example with global adversarial perturbations and $\odot$ denotes Hadamard product. The goal of our Patch Region Localizing is to predict the mask: $\mathbf{F}_{prl}(\mathbf{x}_\emph{adv}) \rightarrow \mathbf{\hat{M}}$ to estimate $\mathbf{M}$.

From the existing attack methods, it can be found that the adversarial patches have two characteristics: the edge of the patch shows strong appearance inconsistency with the neighbor area; the patch placed area has a strong adversarial effect which is also the reason for the missclassification in subsequent model judgment. This adversarial effect is reflected in the feature maps of the high-level layers of the classification model or the detection model, and the position of the adversarial patch appears abnormal large values. In other words, the salient object in the adversarial image is changed to the adversarial patch instead of the original object.  {Therefore, based on the two characteristics, we consider using them for the exploration of adversarial patch areas. Inspired by Multi-task Fully Convolutional Network (MFCN)\cite{salloum2018image}, we consider using edge information to explore the possible patch edges in the image. Then we explore patch areas according to the adversarial effect and synthesize the final predicted patch area. Unlike the MFCN, in which two branches are both pixel-level considerations, our method only uses one branch for edges prediction at the pixel level, and the other branch for prediction based on adversarial effect. Overall, we use a two-branch network model to simultaneously detect the edge lines with appearance inconsistency and the region with abnormal values which are salient.}

For adversarial region localizing, previous studies show that with the increase of the depth of DNNs, the linear calculations and other operations of different layers gradually enlarge the values within the adversarial region. Yu \emph{et al.} \cite{yu2021defending} carry on the mathematical inference and proof. Therefore, we use a deep network as the encoder to extract the features from the image and then construct the first decoder to generate the region directly. For the detection of patch edge lines, LGS \cite{naseer2019local} uses the sobel operator to estimate it by gradient changes which extract all edges in the whole image and cannot distinguish patch regions. Therefore, we adopt the second decoder structure to extract high-level semantic information through residual convolution blocks. By using this semantic information, we detect the patch edge line accurately. Both decoders use the Convolutional Block Attention Module(CBAM) \cite{woo2018cbam}, an attention module for feed-forward convolutional neural networks. Combining channel and spatial dimensions, CBAM uses the intermediate feature map and can effectively help infer the saliency map. 

In practice, we use 8 Resblocks without BatchNorm layers to construct feature extraction network (i.e. the shared encoder). In these 8 Resblock blocks, the number of channels gradually increases from 64 to 256, and features are extracted from multiple angles. For the two branch networks in the PRL module, we use the structure as shown in Table \ref{tab:PRL_REGION}. The CBAM modules can speed up feature extraction and thus accelerate the training process.


\subsubsection{Patch Region Inpainting}
\label{subsec:RI}

After the PRL module detecting the possible patch region, the PRI module uses the predicted $\hat{\bf{M}}$ combined with the input $\mathbf{x}_\emph{adv}$ to recover its original semantic information and eliminate its adversarial perturbations. The PRI module aims to realize a mapping: $\mathbf{F}_\emph{pri}(\mathbf{x}_\emph{adv}| \hat{\bf{M}}) \rightarrow \mathbf{\hat{x}}$ to estimate $\mathbf{x}$.

Intuitively, we hope that the processed image is close to the original image in the appearance. Meanwhile, we hope to eliminate the adversarial attack effect of the patch region in the high-level semantic information extracted by subsequent models. Therefore, we use the output feature maps from the shared encoder and use the decoder structure to reconstruct the patch area based on the extracted semantic information. 

 \begin{table}[tb]
 \centering
 \caption{The structure of the PRL module. } 
 \resizebox{0.48\textwidth}{!}{
 \begin{tabular}{c|c|c|c|c}\hline
 \multicolumn{1}{c|}{Module Name} &
 \multicolumn{1}{c|}{Filter Size}      &  
 \multicolumn{1}{c|}{Channels}   &
 \multicolumn{1}{c|}{Strides/Up Factor}     &
 \multicolumn{1}{c}{Activation}   \\ \cline{1-5}
 \multicolumn{1}{c|}{ResBlock1} & $3\times3$ & 128  & 1    & ELU\\ 
 \multicolumn{1}{c|}{CBAMBlock1} & - & 128  & 1   & ReLU\\ 
 \multicolumn{1}{c|}{NearestUpSample1} &-  & 128  & 2    & -\\\hline 
  \multicolumn{1}{c|}{ResBlock2} & $3\times3$ & 64  & 1    & ELU\\ 
 \multicolumn{1}{c|}{CBAMBlock2} & - & 64  & 1   & ReLU \\\hline  \multicolumn{1}{c|}{ResBlock3} & $3\times3$ & 32  & 1    & ELU\\ 
 \multicolumn{1}{c|}{CBAMBlock3} & - & 32  & 1   & ReLU\\ 
 \multicolumn{1}{c|}{NearestUpSample2} &-  & 32  & 2    & -\\\hline 
 \multicolumn{1}{c|}{ConvBlock1} & $3\times3$ & 16  & 1    & ELU\\ \hline
 \multicolumn{1}{c|}{ConvBlock2} & $3\times3$ & 1 & 1  & Sigmiod\\ \hline
 \end{tabular}
 }
 \label{tab:PRL_REGION}
 \end{table}

In the encoder, we use the variant of the residual network to extract high-level semantic information. However, the patch attack hides the semantic information of the patch region. Although the convolution operation can fill some information by the neighbour area of the patch, this information is not enough to provide strong semantic information in the decoder’s reconstruction. Therefore, we construct a Probabilistic Context Normalization (PCN) blocks \cite{wang2020vcnet} by operation $\mathcal{T(\cdot)}$ to fill in the missing information more effectively. Patch  area and clean area tend to have different domains in distribution. The PCN block uses the mean and variance values of the probabilistic clean area to normalize the patch area, so that the recovered patch area is as identical as possible to the clean area.  {On the basis of PCN, we consider more complex scenarios, especially images with foreground and background scenes, it is inaccurate to reconstruct the patch region using the whole region except the patch part. The distribution of the whole region except the patch part is similar to the whole image, which is very different from the distribution we expect to get for the part to be repaired. Based on this discovery, we redefine the "clean area" and limit it to the region around the patch region.} The improved mathematical definition is as follows:
\begin{equation}
\mathcal{T}(\mathbf{x}_\emph{adv},\hat{\mathbf{M}},\hat{\mathbf{M}}_\emph{d})
={\frac{\mathbf{x}_\emph{m}-\mu\left(\mathbf{x}_\emph{m}\right)}
{\sigma\left(\mathbf{x}_\emph{m}\right)}} 
\cdot \sigma\left(\mathbf{x}_\emph{n}\right)+\mu\left(\mathbf{x}_\emph{n}\right), \label{5}
\end{equation}
\begin{equation}
\mathbf{x}_\emph{n} = \mathbf{x}_{\emph{adv}}\odot(\hat{\mathbf{M}}_\emph{d}-\hat{\mathbf{M}}),  
\mathbf{x}_\emph{m}=\mathbf{x}_{\emph{adv}}\odot\hat{\mathbf{M}}_\emph{d}, \label{6}
\end{equation}
\begin{equation}
\hat{\mathbf{M}}_\emph{d} = {Conv}_\emph{dil}[\hat{\mathbf{M}}],\label{7}
\end{equation}
where $\mu(\cdot)$ and $\sigma(\cdot)$ compute the mean value and the standard deviation of the region respectively. $\hat{\mathbf{M}}_\emph{d}$ denotes the dilated region from $\hat{\mathbf{M}}$ and the kernels of ${Conv}_\emph{dil}$ are $k\times k$ (\emph{k}=3,5,7...) metrics with $dilation > 1$. $(\hat{\mathbf{M}}_\emph{d}-\hat{\mathbf{M}})$ is adopted as we do not need the global semantic information, but only the semantic information around the object where the patch block is located to fit the partial information covered by the patch. In this case, the information of other parts will interfere with the reconstruction of the patch part.  {Subsequent experiments also prove that our considerations are meaningful, and the redefined modules take less time in the training period and the recovered images are of higher quality.}

Similarly, for the PRL module, we use the structure shown in Table \ref{tab:PRI}. Notice that the PRI module is deeper than the PRL module, because the output of the PRI module contains more RGB information rather than the binarization region prediction.
 \begin{table}[tb]
 \centering
 \caption{The structure of the PRI module. } 
 \resizebox{0.48\textwidth}{!}{
 \begin{tabular}{c|c|c|c|c}\hline
 \multicolumn{1}{c|}{Module Name} &
 \multicolumn{1}{c|}{Filter Size}      &  
 \multicolumn{1}{c|}{Channels}   &
 \multicolumn{1}{c|}{Strides/Up Factor}     &
 \multicolumn{1}{c}{Aactivation}   \\ \cline{1-5}
 \multicolumn{1}{c|}{PCNBlock1} & $3\times3$ & 128   & 1    & ELU\\ \hline \multicolumn{1}{c|}{PCNBlock2} & $3\times3$ & 128   & 1    & ELU\\\hline
 \multicolumn{1}{c|}{PCNBlock3} & $3\times3$ & 256  & 1    & ELU\\\hline
 \multicolumn{1}{c|}{PCNBlock4} & $3\times3$ & 256  & 1    & ELU\\\hline
 \multicolumn{1}{c|}{NearestUpSample} & - & 256  & 2    & -\\ \hline
 \multicolumn{1}{c|}{PCNBlock5} & $3\times3$ & 128   & 1    & ELU\\\hline
 \multicolumn{1}{c|}{PCNBlock6} & $3\times3$ & 128   & 1    & ELU\\\hline
 \multicolumn{1}{c|}{PCNBlock7} & $3\times3$ & 64   & 1    & ELU\\\hline
 \multicolumn{1}{c|}{PCNBlock8} & $3\times3$ & 3   & 1    & ELU\\\hline
 
 \multicolumn{1}{c|}{ConvBlock} & $3\times3$ & 3  & 1    & tanh\\ \hline
 \end{tabular}
 }
 \label{tab:PRI}
 \end{table}

\subsubsection{Loss Functions}

It is important to note that the region mask produced by the PRL module will guide subsequent PRI module directly. If the error of the mask value is too large, it will lead the PRI module to focus on the wrong region to mislead the PRI module the semantics of the recovery process. So we train the PRL module independently first and use $\mathcal{L}_\emph{hole}$ loss which is based on $\mathcal{L}_\emph{1}$ loss with weighted parameters to guide the generation of edge and patch area in the PRL: 
\begin{equation}
\begin{split}
\mathcal{L}_{prl}(\hat{\mathbf{x}}, \mathbf{x}) = \mathcal{L}_\emph{prl}((\hat{\mathbf{R}},{\mathbf{R}}),(\hat{\mathbf{E}},{\mathbf{E}})) \\
=\underbrace{\mathcal{L}_\emph{hole}(\hat{\mathbf{R}},{\mathbf{R}})}_{\text {region term }}+\underbrace{\lambda_\emph{edge}\mathcal{L}_\emph{hole}(\hat{\mathbf{E}},{\mathbf{E}})}_{\text {edge term }},\label{3}
\end{split}
\end{equation}
\begin{equation}
\mathcal{L}_\emph{hole}(\hat{\mathbf{I}},{\mathbf{I}})=\emph{relu}(\hat{\mathbf{I}}-\mathbf{I})+\alpha\emph{relu}(\mathbf{I}-\hat{\mathbf{I}}), \label{4}
\end{equation}
where $\hat{\mathbf{E}}$ denotes the outputs of the edge line prediction, and $\mathbf{E}$ denotes the ground truth edge line. $\hat{\mathbf{R}}$ denotes the outputs of the patch region prediction, and $\mathbf{R}$ denotes the ground truth patch region. $\hat{\mathbf{I}}$ denotes the outputs of our generator and $\mathbf{I}$ denotes the ground truth images. $\lambda_\emph{edge}$ denotes the regularization coefficient that balances the influence between edge line and edge region, and $\emph{relu$(\cdot)$}$ is an activation function, and $\alpha$ is a parameter that can be adjusted.

Note that the patch area usually occupies less than 3\% of the whole image, which leads to the hardship of detecting. Simply using $\mathcal{L}_\emph{1}$ loss will make the network tend to generate all black or all white output $\hat{\bf{M}}$, so we adjust the loss weights of the target area and the other area to prevent such outputs and improve the network performance. Experiments also show that the adjusted loss can decrease the training time and avoid falling into all black or white situations.

In order to eliminate the adversarial perturbations of the recovered image, apart from the optimization guidance at the pixel level, we use a pretrained $\emph{VGG19}$ network to extract high-level feature maps and synthesize the guidance information of different feature layers to eliminate its adversarial perturbations: 
\begin{equation}
\mathcal{L}_\emph{fm}(\hat{\mathbf{x}}, \mathbf{x})=
\sum_{i=0}^{n}\beta_\emph{i}\|{V}_\emph{i}(\mathbf{x})-{V}_\emph{i}(\hat{\mathbf{x}})\|_\emph{1},\label{9}
\end{equation}
where ${V}_\emph{i}(\mathbf{x})$ denotes the output of the $i$-$th$ feature map from pretrained $\emph{VGG19}$ model, $\beta_\emph{i}$ is constant, $\emph{n}$ denotes the number of $\emph{VGG19}$ feature maps. Note that since the adversarial perturbations of the image increase with the depth of the network, the amplification of adversarial perturbations in the deep feature map is more obvious.  {The deeper the layers are located, the more obvious the adversarial effect will be, so when eliminating the adversarial effect, we should pay more attention to the features of the high-level. Simply averaging the differences of all layers' features no longer works, so we give larger weight to higher-level features.} In detail, we increase $\beta$ with the depth of the network. 

For the overall network architecture, we use the GAN model. The PRL and PRI parts together form the generator, and the discriminator is composed of the convolution layers and the full connection layer. We use $\mathcal{L}_\emph{gan}$ to denote the GAN loss and use $\mathcal{L}_\emph{g}$ to denote the generator's loss. Therefore, the overall optimization function is:
\begin{equation}
\begin{split}
\mathcal{L}_{\emph{g}}(\hat{\mathbf{x}}, \mathbf{x}) = 
\mathcal{L}_{1}+
\lambda_{fm} \mathcal{L}_\emph{fm}(\hat{\mathbf{x}}, \mathbf{x})\\+\lambda_{gan}\mathcal{L}_\emph{gan}(\hat{\mathbf{x}}, \mathbf{x})+\lambda_{prl}\mathcal{L}_{prl}(\hat{\mathbf{x}}, \mathbf{x}),\label{10} 
\end{split}
\end{equation}
where $\lambda_{fm}$, $\lambda_{prl}, \lambda_{ri}$ and $\lambda_{gan}$ are regularization coefficients to balance each term influence.

\subsubsection{Training Method}

Hope to be able to share an encoder network, we first use the loss of the PRL module and the PRI module to adjust the encoder's parameters, but the effect is not optimal. This result may be caused by the conflict that the inpainting operation requires more semantic information about the image content while the patch region localization network tends to binarize the network feature maps' values. So we finally adopt the following training method:

We initialize the whole network (including the encoder $E$, the PRL module's decoder $D_{PRL}$ and the PRI module's decoder $D_{PRI}$ ) at the beginning.
\begin{enumerate}
\item For the first period, we initialize another encoder $E'$ which has the same structure with the encoder $E$, then train $E'$ and $D_{PRL}$ by the $\mathcal{L}_\emph{hole}$ for few epochs.
\item For the second period, we use the predicted masks from $E'$. Then we train $E$ and $D_{PRI}$ by the loss from the PRI module for a few epochs.
\item For the third period, we fix the parameters in $E$ and fine-tune $D_{PRL}$ by the $\mathcal{L}_\emph{hole}$.
\item For the last period, we fix the parameters in $E$ and $D_{PRL}$, then train the parameters in $D_{PRI}$ by the loss from the PRI module for a few epochs.
\end{enumerate}

After the optimization, we can obtain the optimal encoder and decoder to perform region localizing and inpainting to defend adversarial patch attacks. 
\begin{table*}
\centering
\caption{\small The results of defense performance on different classification models. We report the Top1 Accuracy of the adversarial examples and outputs of different defenses, and the higher value denotes better performance. All our data is from GTSRB \cite{STALLKAMP2012323}. } 
\begin{tabular}{c|c|c|c|c|c|c|c|c|c|c|c|c}
\hline

\multicolumn{1}{c|}{} &
\multicolumn{4}{c|}{{ResNet-50}}      &  
\multicolumn{4}{c|}{{Inception-V3}}   &
\multicolumn{4}{c}{{VGG19} }   \\ \cline{1-13}

\multicolumn{1}{c|} {{Defense}} &
\multicolumn{1}{c|}{Clean}&
\multicolumn{1}{c|}{MS}&  
\multicolumn{1}{c|}{AdvP}&
\multicolumn{1}{c|}{Lavan} & 
\multicolumn{1}{c|}{Clean}&
\multicolumn{1}{c|}{MS}&  
\multicolumn{1}{c|}{AdvP}&
\multicolumn{1}{c|}{Lavan} & 
\multicolumn{1}{c|}{Clean}&
\multicolumn{1}{c|}{MS}&  
\multicolumn{1}{c|}{AdvP}&
\multicolumn{1}{c}{Lavan}  \\ \cline{1-13}
\multicolumn{1}{c|}{No Defense} 
& \textbf{97.37}    & 29.50    & 45.12    & 50.75
& \textbf{90.62}    & 49.57    & 11.87    & 2.13
& \textbf{97.00}    & 35.87    & 19.42    & 22.80
\\ \hline  
\multirow{1}{*}{DW} 
& 90.75    & 53.13    & 80.25    & 80.13 
& 89.12    & 54.83    & 32.13    & 11.11
& 81.87    & 36.55     & 33.96    & 36.09 
\\\hline  
\multirow{1}{*}{LGS} 
& 95.75    & 52.25    & 83.63    & 85.25
& 90.62    & 56.72    & 69.63    & 63.62
& 96.00    & 36.97    & 64.22    & 68.29
\\  \hline  
\multirow{1}{*}{FNC}
& 91.87    & 57.13    & 84.62    & 84.87
& 88.00    & 46.25    & 80.00    & 79.87
& 77.50    & 42.01    & 65.16    & 64.78
\\  \hline  
\multirow{1}{*}{Ours} 
& 95.50    & \textbf{83.88}    & \textbf{92.88}    & \textbf{92.88}
& 90.62   & \textbf{79.88}    & \textbf{90.50}    & \textbf{88.62}
& 96.38   & \textbf{78.25}    & \textbf{89.50}    & \textbf{87.88}
\\ \hline  
\end{tabular}
\label{tab:model}
\end{table*}
\setlength{\tabcolsep}{6pt}

\begin{table*}
\centering
\caption{ The results of attacks on the traffic sign classification task on \emph{ResNet-50}. We report the Top1 Accuracy of the adversarial examples and outputs of different defenses, and the higher value denotes better performance. All data are from GTSRB \cite{STALLKAMP2012323}, and we adopt three patch occupations. } 
\resizebox{0.75\textwidth}{!}{
\begin{tabular}{c|c|c|c|c|c|c|c|c}
\hline
\multicolumn{1}{c|} {Attacks} &
\multicolumn{1}{c|}{Methods} &
\multicolumn{1}{c|}{Clean}&
\multicolumn{1}{c|}{After Attacks}      & 
\multicolumn{1}{c|}{PG }& \multicolumn{1}{c|}{FNC }&
\multicolumn{1}{c|}{LGS }&
\multicolumn{1}{c|}{DW }&
\multicolumn{1}{|c}{\bf{Ours}}  \\ \cline{1-9}

\multirow{3}{*} {MS }
& 2\% &           & 46.07
      & 80.69     & 75.84     
      & 79.57     & 79.25 
      & \bf90.63 \\

& 3\% &  97.37    & 32.00 
      & 74.50     & 63.37     
        &  60.13  &  64.62
        & \bf 87.75  \\
& 5\% &           & 29.50 
      & 69.50     & 57.13     
        & 56.75    & 54.12
        & \bf83.88  \\ \hline
\multirow{3}{*}{AdvP }
& 2\% &           &  68.88
      &  87.13    &  89.88    
        &   90.00    & 85.75
        & \bf 93.38 \\

& 3\% &  97.37    & 20.38 
      &  84.25        &  87.50    
        & 89.00     & 85.00
        & \bf 93.38\\
& 5\% &           & 2.38 
      & 81.00     &  84.12    
        & 82.38   & 79.88
        & \bf 91.25 \\ \hline
\multirow{3}{*}{Lavan} 
& 2\% &           & 70.13
      & 85.38     &  89.63    
      & 90.38    & 87.38
      & \textbf{92.13}  \\

& 3\% &  97.37    &  29.00
      &  84.50    &  88.63   
      &  88.25   & 81.88
      & \textbf{91.00}   \\
& 5\% &           & 2.63
      & 81.88     & 85.25    
      & 85.25    & 67.88
      & \textbf{89.13}  \\ \hline
\end{tabular}
}
\label{tab:clf}
\end{table*}
\section{Experiments and Results}
To demonstrate the effectiveness of our method, we verify its defense performance versus the famous models in computer vision, including classification models and detection models. Because  traffic sign recognition is a core module in autonomous driving, which is a typical security-sensitivity task, we conduct experiments for the traffic sign recognition, i.e., traffic sign classification and traffic sign detection.   We hope that our method can facilitate the robustness improvement of visual recognition in autonomous driving domain.  
\subsection{Experimental Settings}
\begin{table*}[t]
\centering
\caption{\small The results of defense performance on different classification models. We report the SSIM of the adversarial examples and outputs of different defenses, and the higher value denotes better performance. All our data are from GTSRB \cite{STALLKAMP2012323}. } 

\begin{tabular}{c|c|c|c|c|c|c|c|c|c}
\hline
\multicolumn{1}{c|}{} &
\multicolumn{3}{c|}{{ResNet-50}}      &  
\multicolumn{3}{c|}{{Inception-V3}}   &
\multicolumn{3}{c}{{VGG19} }   \\ \cline{1-10}
\multicolumn{1}{c|} {{Defense}} &
\multicolumn{1}{c|}{MS}&  
\multicolumn{1}{c|}{AdvP}&
\multicolumn{1}{c|}{Lavan} & 
\multicolumn{1}{c|}{MS}&  
\multicolumn{1}{c|}{AdvP}&
\multicolumn{1}{c|}{Lavan} & 
\multicolumn{1}{c|}{MS}&  
\multicolumn{1}{c|}{AdvP}&
\multicolumn{1}{c}{Lavan}  \\ \cline{1-10}
\multicolumn{1}{c|}{No Defense} 
& 0.9528   & 0.9354     & 0.9354
& 0.9528    & 0.9353    & 0.9377
& 0.9531    & 0.9385    & 0.9396
\\ \hline  
\multirow{1}{*}{DW} 
& 0.9442    & 0.9272     & 0.9274
& 0.9184    & 0.8973    & 0.8970
& 0.9355    & 0.9284    & 0.9251
\\\hline  
\multirow{1}{*}{LGS} 
& 0.9443    & 0.9346     & 0.9346
& 0.9442    & 0.9352    & 0.9349
& 0.9443    & 0.9376    & 0.9365
\\  \hline  
\multirow{1}{*}{Ours} 
& \textbf{0.9686}    &  \textbf{0.9656}    & \textbf{0.9664}
& \textbf{0.9668}    &  \textbf{0.9458}   & \textbf{0.9490}
& \textbf{0.9672}    & \textbf{0.9514}    & \textbf{0.9516}
\\ \hline  
\end{tabular}
\label{tab:SSIM}
\end{table*}

\begin{table*}[t]
\centering
\caption{The results of defense performance on different classification models. We report the PSNR of the adversarial examples and outputs of different defenses, and the higher value denotes better performance. All our data are from GTSRB \cite{STALLKAMP2012323}. } 

\begin{tabular}{c|c|c|c|c|c|c|c|c|c}
\hline
\multicolumn{1}{c|}{} &
\multicolumn{3}{c|}{{ResNet-50}}      &  
\multicolumn{3}{c|}{{Inception-V3}}   &
\multicolumn{3}{c}{{VGG19} }   \\ \cline{1-10}
\multicolumn{1}{c|} {{Defense}} &
\multicolumn{1}{c|}{MS}&  
\multicolumn{1}{c|}{AdvP}&
\multicolumn{1}{c|}{Lavan} & 
\multicolumn{1}{c|}{MS}&  
\multicolumn{1}{c|}{AdvP}&
\multicolumn{1}{c|}{Lavan} & 
\multicolumn{1}{c|}{MS}&  
\multicolumn{1}{c|}{AdvP}&
\multicolumn{1}{c}{Lavan}  \\ \cline{1-10}
\multicolumn{1}{c|}{No Defense} 
& 21.60    & 17.38     & 17.36
& 21.68    & 18.54    & 18.53
& 21.70    & 18.16     & 17.69
\\ \hline  
\multirow{1}{*}{DW} 
& 21.72    &  18.00    & 18.01
&  21.67   &  18.68   & 18.68
& 21.70    &  18.83   & 18.89
\\\hline  
\multirow{1}{*}{LGS} 
& 21.85    & 21.09     & 21.01
& 21.94    & 20.98    & 20.94
& 21.91    & 20.59    & 20.59
\\  \hline  
\multirow{1}{*}{Ours} 
& \textbf{30.96}    &  \textbf{28.22}    & \textbf{28.66}
& \textbf{30.36}    &  \textbf{23.58}   & \textbf{24.05}
& \textbf{30.89}    & \textbf{26.71}    & \textbf{26.75}
\\ \hline  
\end{tabular}
\label{tab:psnr}
\end{table*}

\noindent{\textbf{Datasets}: To simulate the real situation and generate adversarial patch attacks that can be used in the practical scene, we perform experiments on the open-source datasets: GTSRB \cite{STALLKAMP2012323} and TT100K \cite{Zhe_2016_CVPR}, both are image sets covering a variety of traffic signs which are wildly used in autonomous driving scenes. We use the former dataset to verify our defense on classifiers while using the latter one on object detectors.  {As some images have too few pixels to match the actual image size, we filter the data set. For the classification task, we select 2500 largest images from GTSRB's training set and 800 images from test set images that cover 43 classes. For the object detection task, we also select 546 largest images as training set and 50 images as test set that cover 10 classes. }}

\noindent{\textbf{Threat model}: For classifiers, we choose three models: \emph{Res-Net50} \cite{7780459}, \emph{VGG19} \cite{simonyan2015deep}, \emph{Inception-V3} \cite{7780677}. For object detectors, we choose \emph{Yolov4} \cite{bochkovskiy2020yolov4}. They are all state-of-the-art models as our target models. 
The above classification models obtain 97.37\%, 97.00\% and 90.62\% clean accuracy.}

\noindent{\textbf{Patch Attacks}: Three patch attack methods are adopted: 1) Meaningful Stickers \cite{wei2022adversarial} uses realistic stickers and searches for their locations to perform adversarial attacks. 2) AdvP \cite{brown2017adversarial} and Lavan \cite{karmon2018lavan} randomly choose the patch's locations and then generate the patch's perturbations by different loss functions; For image classifiers, we set patches to occupy 2\%, 3\% and 5\% of the whole image and adopt the above patch attacks. For object detection, we set the patches to occupy 3\% of the entire image.}

\noindent{\textbf{Defenses}: For classifiers, we compare our defense with the above four defenses: Digital Watermarking \cite{hayes2018visible}, Local Gradients Smoothing \cite{naseer2019local}, PatchGuard \cite{xiang2021patchguard} and Feature Norm Clipping \cite{yu2021defending}. For object detectors, as far as we know, there is no targeted defense method against the patch attacks, so we only compare with the preprocessing method LGS. The default parameters are obtained from their papers.} 

\noindent{\textbf{Metrics}: Four metrics: Top1 accuracy, Mean Average Precision (mAP), Structural Similarity (SSIM) and Peak Signal to Noise Ratio (PSNR) are used to evaluate the defense effectiveness. SSIM measures the similarity to the original image, which directly reflects the appearance consistency, and PSNR measures the quality of the output image to reflect whether the defense method improves the prediction accuracy at the expense of image quality.  {Both SSIM and PSNR are used to measure the difference between the output images and the original images. So we use them to evaluate the quality of our output images, and for clean examples, there are no such evaluation.} For all defense methods, we use mAP as the most decisive standard. }

\noindent{\textbf{Implementation}: For the GTSRB dataset, the images $\geq$ 90$\times$90 pixels are selected to perform the experiments because the images under this size are too small to attack. Meanwhile, the TT100K's image sizes are relatively large but the traffic signs in the whole image are relatively small. In order to facilitate subsequent attacks, we crop the local region in which the traffic sign occupies a certain proportion to form the final data set. In addition, as the images have different shapes and sizes, before feeding them into the model, we scale and pad the images to (256$\times$256) and (416$\times$416) for classifiers and object detectors respectively. For the training process, we adopt the batch size of 4 and 2 respectively for the classification task and the object detection task.  {First, we use clean examples to train different kinds of classifiers and detectors. Then, we implement the attacks above. Next, we use the obtained adversarial examples and the corresponding clean examples to train our defense model. Finally, we go through the same process to build the test set and test on the test set. All the attack methods follow the their original paper.}}



\subsection{Evaluate RLID against Patch Attacks}

\subsubsection{Performance Comparisons}

We report the performance of our method against LGS, DW and FNC on different classification models, which shows in Table \ref{tab:model}. For Lavan and AdvP, we take 100 as the maximum number of iterations to train the generic patch to achieve target attacks. And all attacks adopt the patch occupation of 5\%. From the table, we can find that different attacks have different performances under these classification models. Among them, our RLID shows excellence with the accuracy of 87.14\% compared with DW's on 46.46\%, LGS's 64.50\% and FNC's 67.18\% on average. Compared with AdvP and Lavan, the MS's attack success rate is lower but its attack effect is relatively stable. Observing the table, we can find that other defense methods have relatively weak defense capabilities against MS attacks, and our RLID method outperforms other defenses against the MS attack.

We also test our method against LGS, DW, PatchGuard and FNC under different patch occupations. The results of the classification task are shown in Table \ref{tab:clf}. The data shows that examples processed by RLID have an average accuracy of 90.34\% compared with LGS's 80.19\%, DW's 76.19\% PG's 80.98\% and FNC's 80.15\% accuracy in classification tasks. Comparing the defense effects under different patch sizes, it can be found that the advantage of our RLID is obvious under large-size attacks, and it is less affected by the patch size than other defense methods.  {In experiments in Table \ref{tab:clf}, in order to show more situations, we generate one patch for each image instead of a generic patch for all images in the Lavan and the AdvP attacks}. From these data, we can conclude that we achieve outstanding performance in accuracy in classification tasks. 

 {In addition, we explore the defense capability of one defense model trained against another sticker. For example, we train a defense model by the adversarial examples from AdvP and test the defensive effect by the adversarial examples from Lavan. In this situation, RLID can achieve a great defensive effect as the shape of the patch are similar and the adversarial region also can be detected accurately. However, if the shape of the patches is far from the ones in the training set or the patch edges are smoother, the defensive effect will be greatly reduced.}

\subsubsection{Quality Presentation}
\begin{figure*}[t]
\centering
\includegraphics[width=0.95\textwidth]{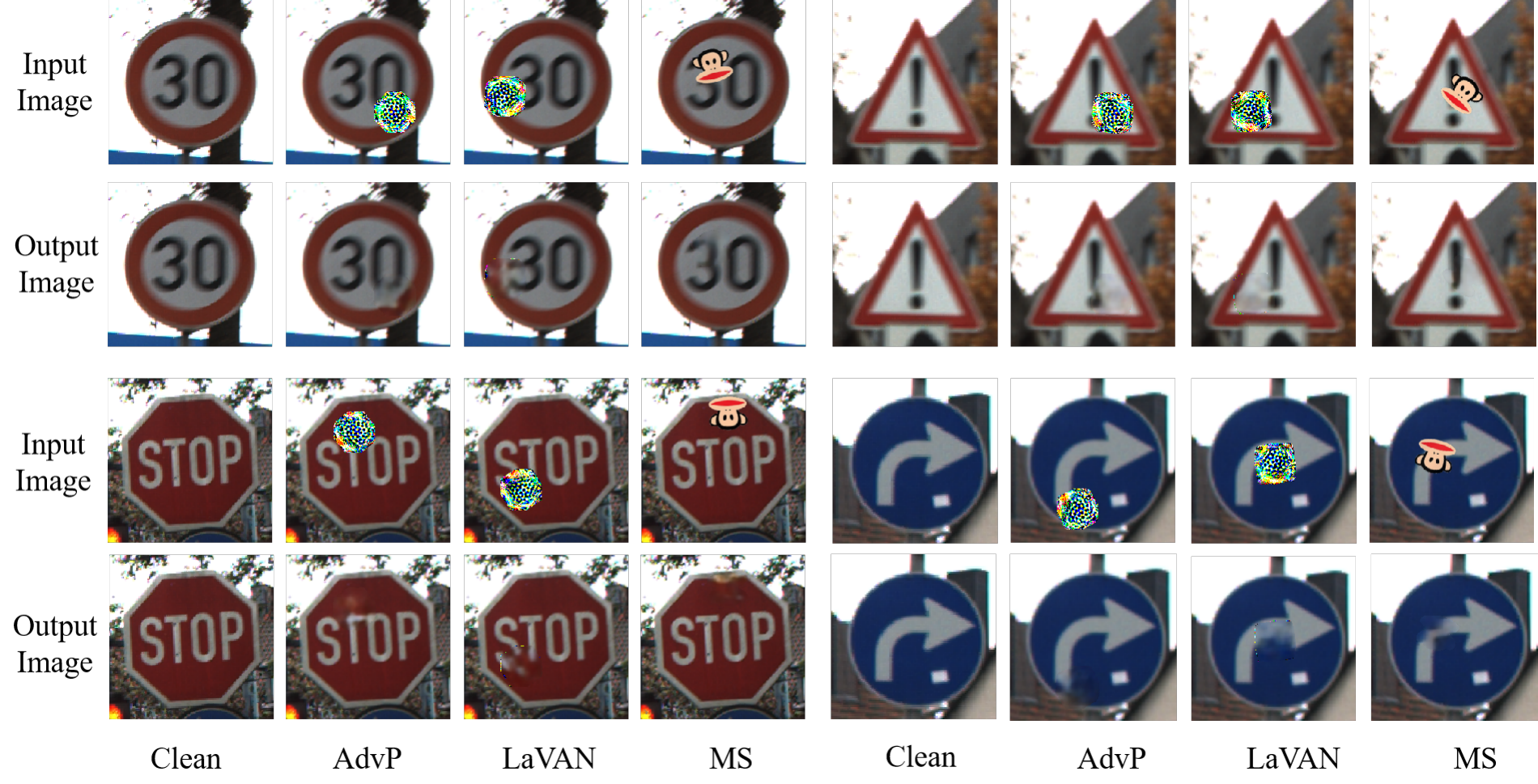}
\caption{There are four groups of examples from our defense methods. In each group, the first row shows the clean image that can be classified correctly and the corresponding adversarial images produced by AdvP, Lavan and MS attacks. The second row shows the recovered images output by our defense methods that can be classified correctly again.}
\label{fig:examples_cls}
\end{figure*}

We report the SSIM and PSNR metrics as Table \ref{tab:SSIM} and Table \ref{tab:psnr} show. Observing Table \ref{tab:SSIM}, we can find that the outputs from DW and LGS perform even worse than the adversarial examples. This is because these two methods cannot distinguish adversarial patch regions from other regions and process the entire image indiscriminately. In contrast, we can see the significant advantage of our RLID. Although its outputs are not the same as the original images, it achieves an average SSIM of 0.9591.

In Table \ref{tab:psnr}, we report PSNR data. 
From the table, we can see that the PSNR values of the adversarial samples and the output pictures of LGS and DW are around 20, while our RLID achieves an average of 27.79, which outperforms other defense methods. The patch occupation of the experimental samples in the above two tables is 5\%.

In addition to qualitative display in the form of metrics, we also present four groups of visual examples in Fig. \ref{fig:examples_cls}. These examples cover different traffic sign types such as speed limits, prohibitions, warnings and indications. It can be seen that the images processed by our method can not only obtain a high classification accuracy, thus effectively defending against the attacks, but also achieve an excellent visual inpainting effect that human eyes almost can not find the difference from the original images.  {As PG and FNC's ``detecting and removing" operation is done in the middle layers of the network, they do not have a final image output, so this section does not display their image quality.}

\subsection{Convergence of RLID}

\begin{figure}
\centering
\includegraphics[width=0.48\textwidth]{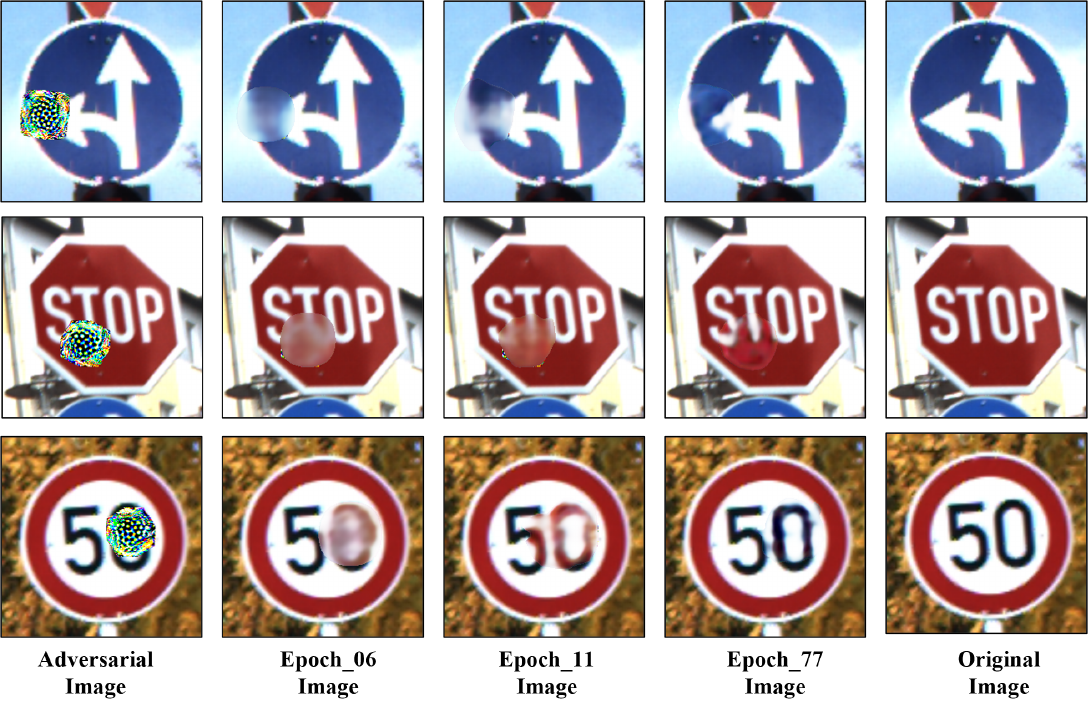}
\caption{Three groups of example in training set from GTSRB \cite{STALLKAMP2012323}. In each row, the leftmost image is the adversarial example, the rightmost image is the original image, and the other images in the middle are RLID outputs after 6, 11 and 77 epochs from left to right respectively.}
\label{fig:epoch}
\end{figure}

\begin{figure}
\centering
\includegraphics[width=0.45\textwidth]{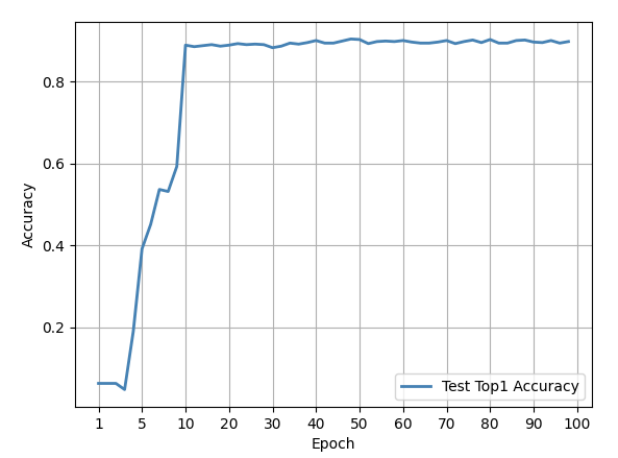}
\caption{The classification accuracy under different training epochs. The classification accuracy reaches 90\% at epoch 11 and slowly converges with fluctuations in subsequent training.}
\label{fig:epoch_line}
\end{figure}

We conduct convergence experiments to verify the convergence of our RLID method and explore the optimal number of epochs that can effectively defend against attacks. Qualitative image presentation and quantitative statistical data are shown in Fig. \ref{fig:epoch} and Fig. \ref{fig:epoch_line}. The experiments are based on the GTSRB data set, and patches occupy 5\% of the entire images in the adversarial examples.

Fig. \ref{fig:epoch} shows the improvement of the image inpainting effect with the increase of the training epochs under the AdvP attack. As can be seen from the figure, the inpainted patch area at the beginning is a blurry patch, and as the iterative training is carried out, the patch area gradually appears textured and has semantic information, and the later training outputs is getting closer and closer to the original image. The visual difference between the final training output and the original image is very slight. The entire training process achieves coarse-to-fine inpainting effect.

Fig. \ref{fig:epoch_line} displays the convergence of classification accuracy as the training epochs increase. From the figure, it can be found that the classification accuracy increases rapidly at the beginning of the training and reaches 90\% in the 11th epoch. In the subsequent training, it increases slowly with fluctuations and gradually converges, reaching the peak of 92\% in the 77th epoch, which shows the good convergence of RLID method.

In terms of image inpainting effect and classification accuracy, although the output of 77 epochs are more similar to the original images visually, the improvement in classification accuracy is only 2\% after more than 60 epochs of training. For overall consideration, in order to achieve effective defense, we can adopt the training model with the best performance in 20 epochs of the training.

\subsection{Defense for Adaptive Attacks}
\begin{table}
\centering
\caption{ The results of BPDA \cite{athalye2018obfuscated} attacks on classification tasks on \emph{ResNet-50}. All data are from GTSRB \cite{STALLKAMP2012323} and the patch occupation in adversarial examples is set to 5\%.} 
\resizebox{0.48\textwidth}{!}{
\begin{tabular}{c|c|c|c|c}
\hline
\multicolumn{1}{c|} {Attacks} &
\multicolumn{1}{c|}{ResNet-50} &
\multicolumn{1}{c|}{ResNet-50+DW}&
\multicolumn{1}{c|}{ResNet-50+LGS}&
\multicolumn{1}{c}{\textbf{ResNet-50+RLID}}  \\ \cline{1-5}
\multirow{1}{*}{AdvP+BPDA}
&   4.38         
         & 5.00  & 77.37
        & \textbf{89.38}  \\ \hline
\multirow{1}{*}{Lavan+BPDA} 
 &  8.75           
        &  15.00   &  68.50 
      & \textbf{88.00}   \\ \hline
\end{tabular}
}
\label{tab:bpda_clf}
\end{table}

The above experiments are based on the assumption that the attackers of various attacks do not know the specific principle of defense method or the exact structure of defense model, but the study \cite{athalye2018obfuscated} points out that adaptive attacks can be realized by taking defense into account through differentiable approximation. It states the phenomena that some defense methods achieve robust defense by implementing obfuscated gradients, a special case of gradient masking. The paper identifies three types of obfuscated gradients and proposes one adaptive attack called BPDA (Backward Pass Differentiable Approximation) to efficiently attack such defense methods. BPDA computes the forward propagation normally and uses a differentiable approximation to compute the backward propagation, thus approximates gradients and bypasses the defense. BPDA also proves its effectiveness against various defense methods through experiments.We also evaluate our RLID method under the BPDA.

We combine BPDA with the AdvP attack and the Lavan attack respectively to attack \emph{ResNet-50} with the 5\% patches. As shown in the Table. \ref{tab:bpda_clf}, we chose LGS and DW to compare the accuracy under the attack. Compared with Table \ref{tab:model}, it can be found that against the attacks with the BPDA attack, the classification accuracy of DW decreases sharply, while LGS, which has better performance, also decreases by about seven percentage points. Our RLID method just declines three percentage points and achieves 89.38\% and 88.00\% respectively, outperforming LGS and DW defense methods. Thus, We can draw the conclusion that the robustness of RLID against adaptive attacks is excellent.

\subsection{Ablation Study}

\begin{figure*}[t]
\centering
\includegraphics[width=0.95\textwidth]{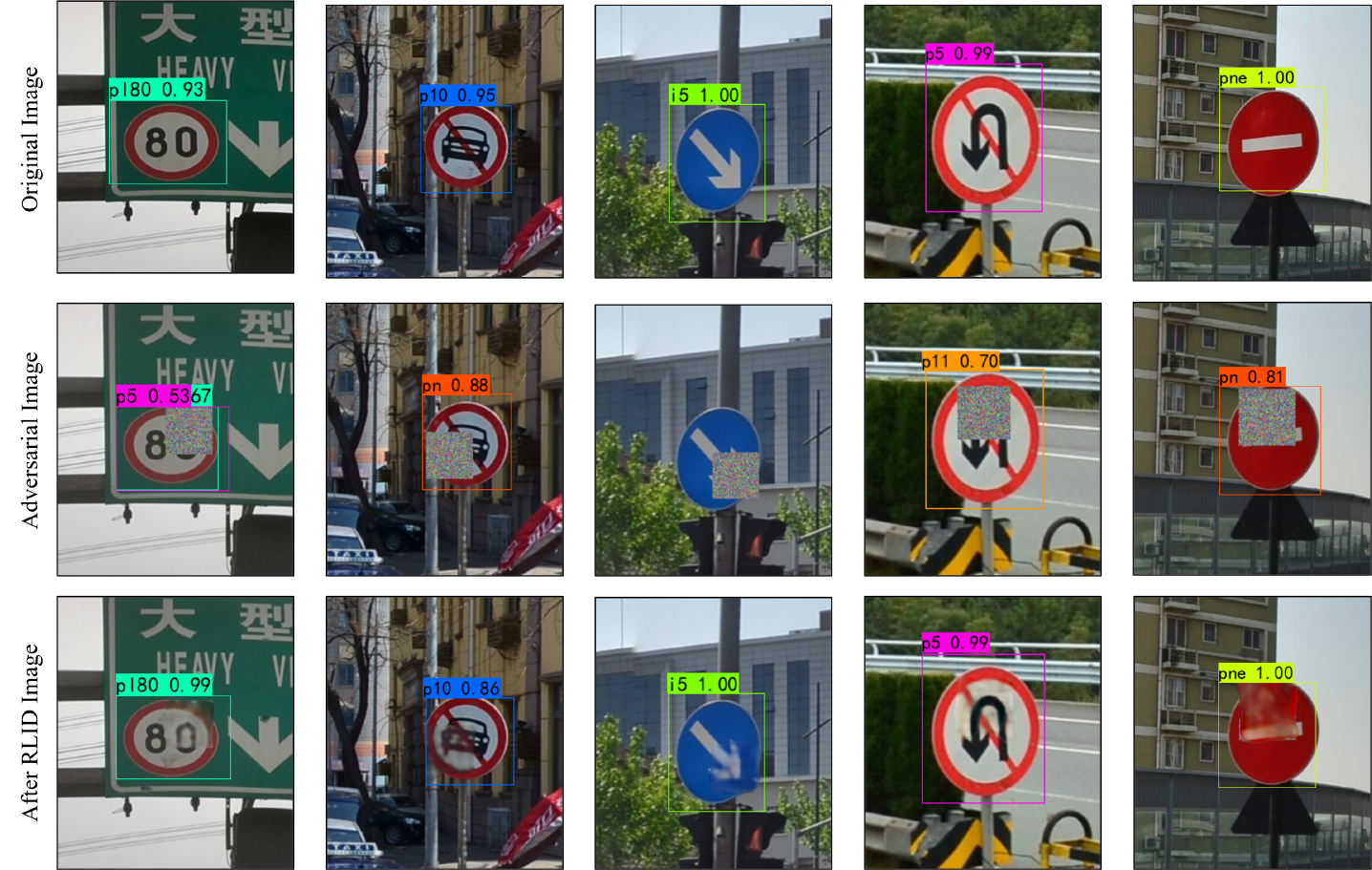}
\caption{There are five groups of examples from our defense methods. In each group, the first row shows the clean image that can be detected correctly and the second row shows the corresponding adversarial images produced by the AdvP attack. The last row shows the outputs of our defense methods that can be detected correctly again.}
\label{fig:examples_obj}
\end{figure*}
\begin{table}
\caption{Ablation study results on GTSRB dataset \cite{STALLKAMP2012323}. All adversarial examples are generated under AdvP attack with the patch occupation of 5\%. The training is based on the same data set and the same training method.}
\centering  
\begin{tabular}{c|c|c|c}
\hline
\multicolumn{1}{c|} {}    &{Accuracy} & {PSNR} & {SSIM}           \\ \cline{1-4}
              RLID w/o inpainting &    73.88\%   & 18.13     & 0.8760          \\\hline
RLID w/o dilation   &    87.38\%   &  25.98    & 0.9494                           \\ \hline
RLID w/o region     &    91.75\%   & 27.89     & 0.9560                                          \\ \hline
full-version RLID      &  \textbf{92.88\%}     &     \textbf{28.22}  & \textbf{0.9656}        \\ \hline
\end{tabular}

\label{tab:ablation}
\end{table}
To demonstrate the effectiveness of each component in the proposed method, we report the performance without each component of the network on classification tasks. We conduct experiments under three attacks, the results are similar, and we use the AdvP's result. In the experiment, we used adversarial samples for \emph{ResNet-50} and the results are shown in Table \ref{tab:ablation}. After the attack, the classification task's accuracy falls to 45.12\%. We uses the same training strategy to train on these adversarial samples. 

\noindent{\textbf{``RLID w/o inpainting''}: indicates that after the patch area is located, it will be filled with a pure color (we use black in the experiment), and this experiment is used to prove the importance of the PRI module.}

\noindent{\textbf{``RLID w/o dilation''}: indicates that the initial filling pixels of the patch area is carried out by using the entire image information instead of the adjacent pixels' information in the PRI module, which was described in detail in the previous section \ref{subsec:RI}.}

\noindent{\textbf{``RLID w/o region''}: indicates that only use the results of the edge prediction module as the final results of patch area detection in PRL module. Note that since the results of the region prediction tend to take on the shape of edge lines, it is also necessary to fill the edges before using, as mentioned in the previous section \ref{subsec:prl}.}


From the Table \ref{tab:ablation}, it can be seen that under the three metrics, the values without the PRI module decreased significantly compared with the full-version RLID (accuracy decreases 20 percentage points, PSNR decreases 10, SSIM decreases 0.09). Secondly, when the dilation operation is not used, each metric data will also have a decline which can not be ignored. Finally, it can be found that both branches are not as effective when used separately as when used together. Based on the above experiments, we can conclude that the most important parts of RLID is the PRI module, and the dilation operation also plays a role. At the same time, using the two-branch structure can achieve the optimal effect.

\subsection{RLID in Object Detection Tasks}

\begin{table}
\centering
\caption{\small The results of attacks on the traffic sign detection task on \emph{Yolov4} \cite{bochkovskiy2020yolov4}. We report the Mean Average Precision (mAP) of the adversarial examples and outputs of different defenses and the higher value denotes better performance. All data are from TT100k dataset \cite{Zhe_2016_CVPR}. } 
\begin{tabular}{c|c|c|c|c}
\hline
\multicolumn{1}{c|} {Attacks}  &  \multicolumn{1}{c|}{Clean}&   \multicolumn{1}{c|}{After Attack}      &  \multicolumn{1}{c|}{LGS }&  \multicolumn{1}{c}{\bf{Ours}} \\ \cline{1-5}
\multirow{1}{*}{MS }        & 0.89           & 0.18          & 0.07          &\bf{0.51}  \\  \hline
\multirow{1}{*}{AdvP }    & 0.89         & 0.22          & 0.27         & \bf{0.44}  \\ \hline
\multirow{1}{*}{Lavan }    & 0.89              & 0.25            & 0.10          &\bf{0.45}    \\ \hline
\end{tabular}
\label{tab:ob_detect}
\end{table}

\subsubsection{Performance Comparisons}
We report the performance of our method against LGS in the object detection task. Results are shown in Table \ref{tab:ob_detect}. The average mAP after RLID is 0.47 compared with LGS's 0.15 in object detection tasks.  {Although the results fall compared to the clean sample, the defensive effect of RILD is significantly improved compared to the other defense method.} 


\subsubsection{Quality Presentation}
Five groups of visual examples are presented in Fig. \ref{fig:examples_obj}. These examples also cover different traffic sign types. It can be seen that after attack, the object detector may fail to detect objects or misclassify objects in the images. After RLID, the images processed by our method can not only obtain a high classification accuracy, thus effectively defending against the attacks, but also achieve an excellent visual inpainting effect.  {However, when compared to inpainting images in the classification task, we can see that the inpainting effect is slightly less effective. According to our analysis, on the one hand, images from object detection tasks are larger than those from classification tasks and larger images bring more difficulty to repair. More fine-grained information is needed. On the other hand, there is more interference from background areas which also brings more difficulty to repair. In summary, RLID's defensive effect will be weakened to large images with small targets.}

\section{Conclusion}

We proposed Region Localizing and Inpainting Defense (RLID), a novel defense method against patch attacks in real-world scenes. Our method first detected the patch's location and then used the localization result to inpaint the image. Combining with the consideration of pixel-level inconsistency and adversarial effect, we used two-branch learning to localize the patch region and used the loss function based on high-level feature maps to inpaint the adversarial examples. We trained the defense network via an iterative optimization manner and conducted a series of experiments to implement the state-of-the-art adversarial patch attacks and state-of-the-art defenses to these attacks. We achieved the average recovering success rate of 57\%, outperforming the other defenses. The restored images also had high picture quality. We believe this paper can provide some insights to the defense for adversarial patches.











\printcredits

\

\bibliographystyle{model1-num-names}

\bibliography{cas-dc-template}


\end{document}